\documentclass{llncs}
\usepackage{graphicx}
\usepackage{caption}
\usepackage{multirow}
\usepackage{booktabs}

\begin{document}

\title{Joint Learning of Motion Estimation and Segmentation for Cardiac MR Image Sequences}

\author{Chen Qin\inst{1}, Wenjia Bai\inst{1}, Jo Schlemper\inst{1}, Steffen E. Petersen\inst{2}, Stefan K. Piechnik\inst{3}, Stefan Neubauer\inst{3}, and Daniel Rueckert\inst{1}
}

\institute{Department of Computing, Imperial College London, London, UK\\
\and
NIHR Biomedical Research Centre at Barts, Queen Mary University of London, London, UK\\
\and Division of Cardiovascular Medicine, Radcliffe Department of Medicine, University of Oxford, Oxford, UK\\}

\maketitle              

\begin{abstract}
Cardiac motion estimation and segmentation play important roles in quantitatively assessing cardiac function and diagnosing cardiovascular diseases. In this paper, we propose a novel deep learning method for joint estimation of motion and segmentation from cardiac MR image sequences. The proposed network consists of two branches: a cardiac motion estimation branch which is built on a novel unsupervised Siamese style recurrent spatial transformer network, and a cardiac segmentation branch that is based on a fully convolutional network. In particular, a joint multi-scale feature encoder is learned by optimizing the segmentation branch and the motion estimation branch simultaneously. This enables the weakly-supervised segmentation by taking advantage of features that are unsupervisedly learned in the motion estimation branch from a large amount of unannotated data. Experimental results using cardiac MRI images from 220 subjects  show that the joint learning of both tasks is complementary and the proposed models outperform the competing methods significantly in terms of accuracy and speed. 

\end{abstract}
\section{Introduction}
Cardiac magnetic resonance imaging (MRI) is one of the reference methods to provide qualitative and quantitative information of the morphology and function of the heart, which can be utilized to assess cardiovascular diseases. Both cardiac MR image segmentation and motion estimation are crucial steps for the dynamic exploration of the cardiac function, which enable the accurate quantification of regional function measures such as changes in ventricular volumes and the elasticity and contractility properties of the myocardium \cite{shen2005consistent}. Traditionally, most approaches consider segmentation and motion estimation as two separate problems. 
However, these two tasks are known to be closely related \cite{cheng2017segflow,tsai2016video}, and learning meaningful representations for one problem should be helpful to learn representations for the other one.

In this paper, we propose a joint deep learning network for predicting the segmentation and motion estimation simultaneously for cardiac MR sequences. In particular, the proposed architecture consists of two branches: one is an unsupervised Siamese style spatial transformer network for cardiac motion estimation, which exploits multi-scale features and recurrent units to accurately predict sequences of motion fields while ensuring spatio-temporal smoothness; and the other one is a segmentation branch which takes advantage of the joint feature learning to enable weakly-supervised segmentation for temporally sparse annotated data. We formulate the problem as a composite loss function optimized by training both tasks simultaneously. Using experiments with cardiac MRI from 220 subjects, we show that the proposed models can significantly improve the performance.

\subsection{Related Work}
In recent years, many works in deep learning domain have been proposed for cardiac MR image segmentation. Most of these approaches employ a fully convolutional network which learns useful features by training on manually annotated images and predicts a pixel-wise label map \cite{avendi2016combined,bai2017semi,bai2017human,ngo2017combining}. However, in real world applications, normally only end-distolic (ED) and end-systolic (ES) frames are manually annotated in a sequence of cardiac MR images, while information contained in other frames is not exploited in previous works. On the other hand, traditional methods commonly extended classical optical flow or image registration methods for cardiac motion estimation \cite{de2012temporal,shen2005consistent,shi2012comprehensive,tobon2013benchmarking}. For instance, De Craene et al. \cite{de2012temporal}  optimized a 4D velocity field parameterized by B-Spline spatio-temporal kernels to introduce temporal consistency,  and Shi et al. \cite{shi2012comprehensive} combined different MR sequences to estimate myocardial motion using a series of free-form deformations (FFD) \cite{rueckert1999nonrigid}. In recent years, some deep learning works \cite{simonovsky2016deep,uzunova2017training} have also been proposed for medical image registration. They either trained networks to learn similarity metrics or simulated transformations as ground truth to learn the regression. In contrast, our proposed method is a unified model for learning both cardiac motion estimation and segmentation, where no motion ground truth is required and only temporally sparse annotated frames in a cardiac cycle are needed. Of particular relevance to our approach are works \cite{cheng2017segflow,patraucean2015spatio} proposed in computer vision domain. Segflow  \cite{cheng2017segflow} used a joint learning framework for natural video object segmentation and optical flow, and the work \cite{patraucean2015spatio} propagated labels using the estimated flow to enable weakly-supervised segmentation. In comparison, our method proposes a different way to couple both tasks by learning a joint feature encoder, which exploits the massive information contained in unlabeled data and explores the redundancy of the feature representation for both tasks. 

\section{Methods}
\begin{figure*}[!t]
\centering
\includegraphics[width=\linewidth]{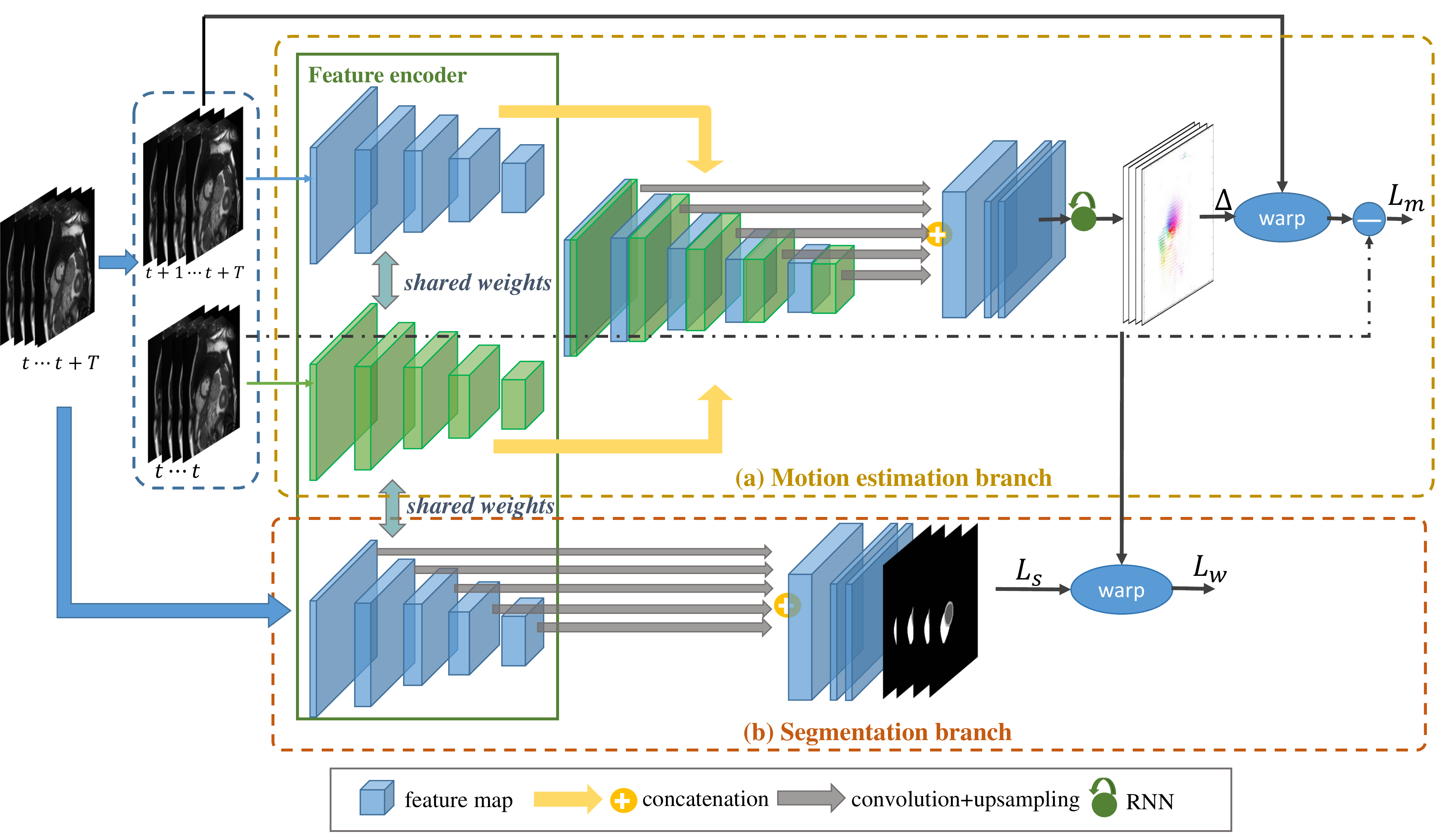}
\caption{The overall schematic architecture of proposed network for joint estimation of cardiac motion and segmentation. (a) The proposed Siamese style multi-scale recurrent motion estimation branch. (b) The segmentation branch which shares the joint feature encoder with motion estimation branch. The architecture for feature encoder is adopted from VGG-16 net before FC layer. Both branches have the same head architecture as the one proposed in \cite{bai2017human}, and the concatenation layers of motion estimation branch are from last layers at different scales of the feature encoder.}
\label{fig:joint_model}
\end{figure*}

Our goal is to realize the simultaneous motion estimation and segmentation for cardiac MR image sequences. Here we construct a unified model consisting of two branches: an unsupervised motion estimation branch based on a Siamese style recurrent multi-scale spatial transformer network, and a segmentation branch based on a fully convolutional neural network, where the two branches share a joint feature encoder. The overall architecture of the model is shown in Fig. \ref{fig:joint_model}.
\subsection{Unsupervised Cardiac Motion Estimation}
\label{motion estimation}
Deep learning methods normally rely heavily on the ground truth labeled data. However, in problems of cardiac motion estimation, dense transformation maps between frames are rarely available. Inspired by the success of spatial transformer network \cite{caballero2017real,jaderberg2015spatial,patraucean2015spatio} which effectively encodes optical flow to describe motion, here we propose a novel Siamese style multi-scale recurrent network for estimating the cardiac motion of MR image sequences without supervision effort. A schematic illustration of the model is shown in Fig. \ref{fig:joint_model}(a).

The task is to find a sequence of consecutive optical flow representations between the target frame $I_{t}$ and the source frames $I_{t+1}, I_{t+2}, ..., I_{t+T}$, where the output is pixel-wise 2D motion fields $\Delta$ representing the displacement in $x$ and $y$ directions. In order to realize this,  the proposed network mainly consists of four components: a Siamese network for the feature extraction of both target frame and source frame; a multi-scale concatenation of features from pairs of frames; a convolutional recurrent unit (RNN) which propagates information along temporal dimension; and a sampler that warps the source frame to the target one by using the estimated  displacement field. In details, inspired by the success of cardiac segmentation network proposed in \cite{bai2017human}, we determine the Siamese feature encoder as the one in \cite{bai2017human} which is adapted from VGG-16 net. For the combination of information from frame pairs, motivated by the traditional multi-level registration method \cite{rueckert1999nonrigid}, here we propose to concatenate multi-scale features from both streams of Siamese network to exploit information at different scales. This is followed by a convolution and upsampling operation back to the original resolution, and combined using a concatenation layer. In addition, in order to exploit information from consecutive frames and also to ensure the spatio-temporal smoothness of the estimated motion fields, we additionally incorporate a convolutional simple RNN with tanh function at the last layer to propagate motion information along the temporal dimension and to estimate flow with two feature maps $\Delta = (\Delta x, \Delta y; \theta_{\Delta})$ corresponding to displacements for the $x$ and $y$ dimensions, where the network is parameterized by $\theta_{\Delta}$. Finally,  the source frames $I_{t+k}$ are transformed using bilinear interpolation to the target frame, which can be expressed as $I_{t+k}^{'}(x,y) = \Gamma\{I_{t+k}(x+\Delta_{t+k}x, y+\Delta_{t+k}y)\}$. 

To train the spatial transformer, we optimize the network by minimizing the pixel-wise mean squared error between the transformed frames and the target frame. To ensure local smoothness, we penalize the gradients of flow map by using an approximation of Huber loss proposed in \cite{caballero2017real}, namely $\mathcal{H}(\delta_{x,y}\Delta_t) = \sqrt{\epsilon+\sum_{i=x,y}(\delta_{x}\Delta i^2+\delta_{y}\Delta i^2)}$  and similarly, we use a regularization term $\mathcal{H}(\delta_{t}\Delta) = \sqrt{\epsilon+\sum_{i=x,y,t}\delta_{t}\Delta i^2}$  to constrain the flow to behave smoothly in temporal dimension, where $\epsilon=0.01$.  Therefore, the loss function can be described as follows:
\begin{equation}
\mathcal{L}_m = \frac{1}{T}\sum_{k=1}^{T}[\|I_t-I_{t+k}^{'}\|^2+\alpha \mathcal{H}(\delta_{x,y}\Delta_{t+k})] + {\beta} \mathcal{H}(\delta_{t}\Delta),
\end{equation}
where $T$ is the number of sequence,  $\alpha$ and $\beta$ are regularization parameters to trade off between image dissimilarity, local and temporal smoothness.

\subsection{Joint Model for Cardiac Motion Estimation and Segmentation}
\label{joint model}
 As we know, motion estimation and segmentation tasks are closely related, and previous works in computer vision domain have shown that the learning of one task is able to benefit the other \cite{cheng2017segflow,tsai2016video}.  Motivated by the success of self-supervised learning which learns features from intrinsic freely available signals \cite{agrawal2015learning,doersch2017multi}, here we propose a joint learning model for cardiac motion estimation and segmentation, where features learned from unsupervised (or self-supervised) motion estimation are exploited for segmentation.   By coupling the motion estimation and segmentation network, the proposed approach can be viewed as a weakly-supervised method with temporally sparse annotated data while motion estimation facilitates the feature learning by exploring those unlabeled data. The schematic architecture of the unified model is shown in Fig. \ref{fig:joint_model}.

In details, the proposed joint model consists of two branches: the motion estimation branch proposed in Section \ref{motion estimation}, and the segmentation branch based on the effective network proposed in \cite{bai2017human}. Here both branches share the joint feature encoder (Siamese style network) as shown in Fig. \ref{fig:joint_model}, so that the features learned can better capture the useful related representations for both tasks. 
Here a categorical cross-entropy loss $\mathcal{L}_s = -\sum_{l\in L} y_{l} \textup{log}( f(x_{l};\Theta) )$ on labeled data set $L$ is used for segmentation branch, in which we define $x_l$ as the input data, $y_l$ as the ground truth, and $f$ is the segmentation function parameterized by $\Theta$. In addition, to further exploit the input unlabeled data, we add an additional spatial transformer in segmentation branch, which warps the predicted segmentation to the target frame using the motion fields estimated from motion estimation branch. Similarly, a categorical cross-entropy loss $\mathcal{L}_w = -\sum_{n\in U} y_{l} \textup{log}( f_w(x_{n};\Theta) )$ is used between the warped segmentations and the target, where $U$ stands for unlabeled data set, and $f_w$ is $f$ plus the warp operation. This component mainly works as a regularization for the motion estimation branch, which is supposed to improve the estimation around boundaries.

As a result, a composite loss function consisting of image similarity error, smoothness penalty of motion fields, and pixel-wise cross entropy segmentation losses with the softmax function can be defined as follows:
\begin{equation}
\mathcal{L} = \mathcal{L}_{m}+\lambda_1\mathcal{L}_{s}+\lambda_2 \mathcal{L}_{w},
\end{equation}
where $\lambda_1$ and $\lambda_2$ are trade-off parameters for different tasks. To initialize the joint model, we first train the motion estimation branch using all the available data we have. Then we fix the weights of the shared feature encoder, and train the segmentation branch with the available annotated data. Lastly, we jointly train both branches by minimizing the composite loss function on training set.

\section{Experiments and Results}
Experiments were performed on 220 short-axis cardiac MR sequences from UK Biobank study. Each scan contains a sequence of 50 frames, where manual segmentations of left-ventricular (LV) cavity, the myocardium (Myo) and the right-ventricular (RV) cavity are available on ED and ES frames. A short-axis image stack typically consists of 10 image slices. For pre-processing, all training images were cropped to the same size of $192\times192$,  and intensity was normalized to the range of [0,1]. In our experiments, we split the data into 100/100/20 for training/testing/validation. Parameters used in the loss function were set to be  $\alpha = 0.001$, $\beta = 0.0001$,  $\lambda_1=0.01$ and $\lambda_2=0.001$, which were chosen via validation set. The number of image sequence for RNN during training was $T=10$, and a learning rate of 0.0001 was used. Data augmentation was performed on-the-fly, with random rotation, translation, and scaling.

Evaluation was performed with respect to both segmentation and motion estimation. We first evaluated the segmentation performance of the joint model by comparing it with the baseline method, i.e., training the segmentation branch only (Seg only). Results reported in Table \ref{segmentaton_evaluation} are Dice scores computed with manual annotations on LV, Myo, and RV. It shows that the proposed joint model significantly outperforms the results of Seg only on all three structures with $p \ll 0.001$ using Wilcoxon signed rank test, especially on Myo where motion normally affects the segmentation accuracy greatly. This indicates the merits of joint feature learning, where features explored by motion estimation are beneficial for segmentation task. 
\begin{table*}[!t]
  \centering
  \caption{Evaluation of segmentation accuracy for the proposed joint model and the baseline (Seg only) method in terms of the Dice Metric (mean and standard deviation).}
  \label{segmentaton_evaluation}
   \setlength{\tabcolsep}{5.5pt}
  \begin{tabular}{cccc}  
    \toprule
  {Method} & {LV} & {Myo}   & RV     \\
  \midrule
     {Seg only} & 0.9217 (0.0450) & 0.8373 (0.0309)  &  0.8676 (0.0513) \\
 Joint model & \textbf{0.9348 (0.0408)} & \textbf{0.8640 (0.0295)} & \textbf{0.8861 (0.0453)}   \\
    \bottomrule
  \end{tabular}
\end{table*}

We also evaluated the performance of motion estimation by comparing the results obtained using a B-spline free-form deformation (FFD) algorithm\footnote{https://github.com/BioMedIA/MIRTK} \cite{rueckert1999nonrigid}, network proposed in Section \ref{motion estimation} (Motion only), and the joint model proposed in Section  \ref{joint model}. We warped the segmentations of ES frame to ED frame by using the estimated motion fields, and mean contour distance (MCD) and Hausdorff distance (HD) were computed between the transformed  segmentations and the segmentations of ED frame. Table \ref{motion_evaluation} shows the comparison results of these methods. It can be observed that both of the proposed methods outperform FFD registration method in terms of MCD and HD on all the three structures ($p \ll 0.001$) and similarly, the joint model shows better performance than the model trained for motion estimation only ($p \ll 0.001$ on LV and RV, and $p<0.01$ on Myo). Additionally, we compared the test time needed for motion estimation on 50 frames of a single slice in a cardiac cycle, and results indicated a faster speed of proposed methods compared to FFD. 
\begin{table*}[!t]
  \centering
  \caption{Evaluation of motion estimation accuracy for FFD, proposed model in Section \ref{motion estimation} (Motion only) and the proposed joint model in terms of the mean contour distance (MCD) and Hausdorff distance (HD) in mm (mean and standard deviation). Time reported is testing time on 50 frames in a cardiac cycle per slice.}
  \label{motion_evaluation}
\scalebox{0.95}
 { \begin{tabular}{cccccccc}
  
    \toprule
\multirow{2}*{Method} & \multicolumn{3}{c}{MCD} & \multicolumn{3}{c}{HD}& \multirow{2}*{Time}\\
\cline{2-7}
 & {LV} & {Myo}   & RV  & {LV} & {Myo}   & RV     \\
  \midrule
  {FFD} & 1.83(0.53)  & 2.47(0.74) &  3.53(1.25) & 5.10(1.28) & 6.47(1.69) & 12.04(4.85) & 13.22s\\
 {Motion only} & 1.55(0.49) & 1.23(0.30) & 3.14(1.12) & 4.20(1.04) & 3.51(0.88) & 11.72(\textbf{4.28}) & \textbf{1.23}s\\
 Joint model & \textbf{1.30}(\textbf{0.34}) & \textbf{1.19}(\textbf{0.26}) & \textbf{3.03}(\textbf{1.08})  & \textbf{3.52}(\textbf{0.82}) & \textbf{3.43}(\textbf{0.87}) & \textbf{11.38}(4.34) &{2.80}s\\
    \bottomrule
  \end{tabular}}
\end{table*}

\begin{figure*}[!t]
\centering
\includegraphics[width=\linewidth]{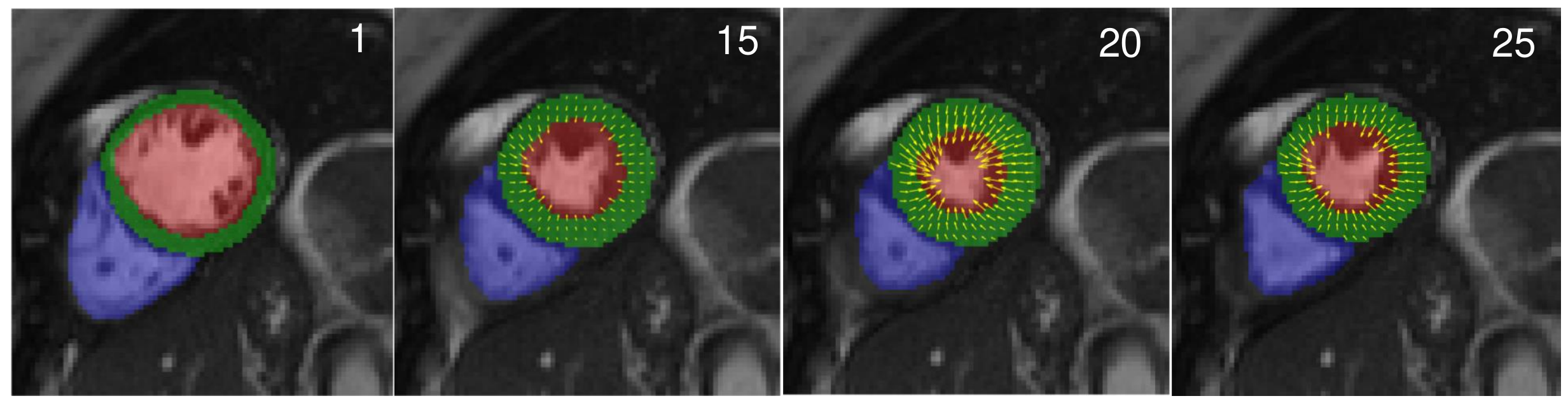}
\caption{Visualization results for simultaneous prediction of motion estimation and segmentation. Myocardial motions are from ED to other time points. Please refer to supplementary material for a dynamic video of a cardiac cycle.}
\label{fig:visualization}
\end{figure*}

\begin{figure*}[!t]
\centering
\includegraphics[width=0.85\linewidth]{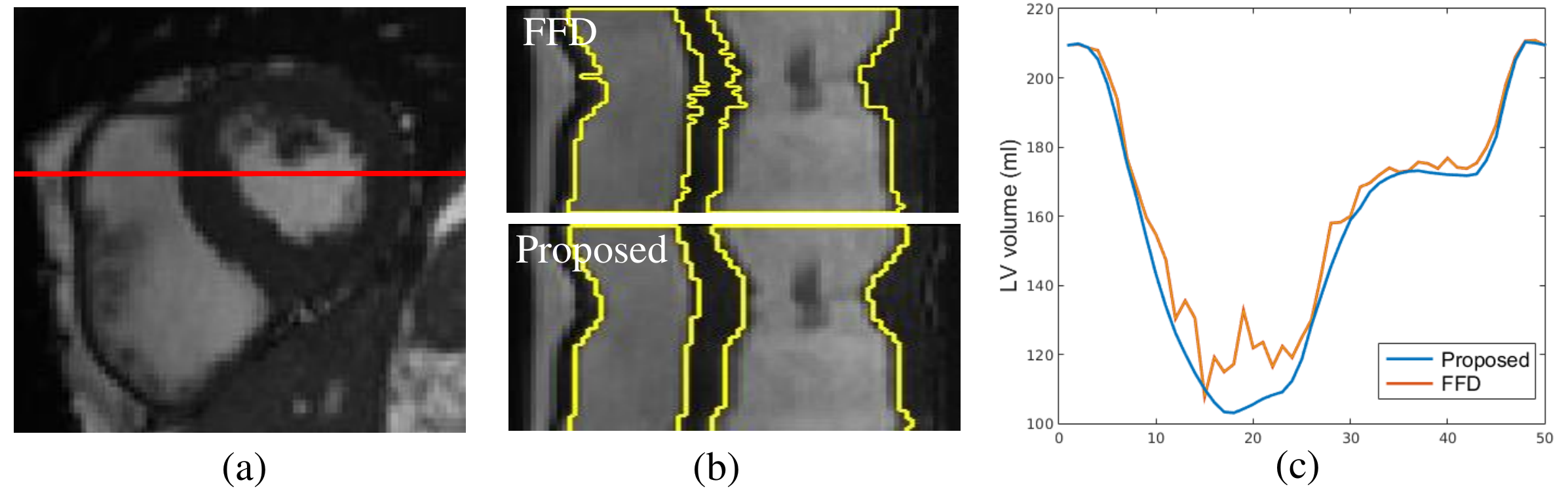}
\caption{(a) (b)Labeling results obtained by warping the ED frame segmentation to other time points using FFD and the proposed joint model. Results are shown in temporal views of the red short-axis line. (c) Left ventricular volume (ml) of the subject by warping the ED frame segmentation to other time points in a cardiac cycle.}
\label{fig:temporal_smooth}
\end{figure*}

Furthermore, the proposed joint method is capable of predicting a sequence of estimated motion fields and segmentations simultaneously. Here we show a visualization result of the network predictions with segmentations and motions combined on frames in a cardiac cycle in Fig. \ref{fig:visualization}. Myocardial motion indicated by the yellow arrows were established between ED and other time frames. Note that the network predicts dense motion fields, while for better visualization, we only show a sparse representation around myocardium. To further validate the proposed unified model in terms of the motion estimation, Fig. \ref{fig:temporal_smooth}(a)(b) shows a labeling results of the LV and RV boundaries along temporal dimension, which is obtained by warping the labeled segmentations available in ED frame to other time points, and  Fig. \ref{fig:temporal_smooth}(c) calculated the transformed LV volume over the cardiac cycle. These show that the proposed model is able to produce an accurate estimation, which is also smooth and consistent over time. 

\section{Conclusion}
In this paper, we have presented a novel deep learning model for joint motion estimation and segmentation of cardiac MR image sequence. The proposed architecture is composed of two branches: a proposed unsupervised Siamese style recurrent spatial transformer network for motion estimation and  a segmentation branch based on a fully convolutional network. A joint feature encoder is shared between the two branches, which enables  the effective feature learning via multi-task training and also the weakly-supervised segmentation in terms of the temporally sparse annotated data. Experimental results showed significant improvements of proposed models against baseline approaches in terms of accuracy and speed.
For the future work, we will validate our method on a larger scale dataset, and will also investigate  its usefulness on 3D applications.

\bibliographystyle{splncs03}
\bibliography{ref}

\begin{thebibliography}{10}
\providecommand{\url}[1]{\texttt{#1}}
\providecommand{\urlprefix}{URL }

\bibitem{agrawal2015learning}
Agrawal, P., Carreira, J., Malik, J.: Learning to see by moving. In: ICCV. pp.
  37--45 (2015)

\bibitem{avendi2016combined}
Avendi, M., Kheradvar, A., Jafarkhani, H.: A combined deep-learning and
  deformable-model approach to fully automatic segmentation of the left
  ventricle in cardiac {MRI}. Medical image analysis  30,  108--119 (2016)

\bibitem{bai2017semi}
Bai, W., Oktay, O., Sinclair, M., et~al.: Semi-supervised learning for
  network-based cardiac {MR} image segmentation. In: MICCAI. pp. 253--260
  (2017)

\bibitem{bai2017human}
Bai, W., Sinclair, M., Tarroni, G., et~al.: Automated cardiovascular magnetic
  resonance image analysis with fully convolutional networks. Journal of
  Cardiovascular Magnetic Resonance  (2018)

\bibitem{caballero2017real}
Caballero, J., Ledig, C., Aitken, A., et~al.: Real-time video super-resolution
  with spatio-temporal networks and motion compensation. In: CVPR (2017)

\bibitem{cheng2017segflow}
Cheng, J., Tsai, Y.H., Wang, S., Yang, M.H.: Segflow: Joint learning for video
  object segmentation and optical flow. In: ICCV. pp. 686--695 (2017)

\bibitem{de2012temporal}
De~Craene, M., Piella, G., Camara, O., et~al.: Temporal diffeomorphic free-form
  deformation: Application to motion and strain estimation from {3D}
  echocardiography. Medical image analysis  16(2),  427--450 (2012)

\bibitem{doersch2017multi}
Doersch, C., Zisserman, A.: Multi-task self-supervised visual learning. In:
  ICCV (2017)

\bibitem{jaderberg2015spatial}
Jaderberg, M., Simonyan, K., Zisserman, A., et~al.: Spatial transformer
  networks. In: NIPS. pp. 2017--2025 (2015)

\bibitem{ngo2017combining}
Ngo, T.A., Lu, Z., Carneiro, G.: Combining deep learning and level set for the
  automated segmentation of the left ventricle of the heart from cardiac cine
  magnetic resonance. Medical image analysis  35,  159--171 (2017)

\bibitem{patraucean2015spatio}
Patraucean, V., Handa, A., Cipolla, R.: Spatio-temporal video autoencoder with
  differentiable memory. ICLR workshop  (2016)

\bibitem{rueckert1999nonrigid}
Rueckert, D., Sonoda, L.I., Hayes, C., et~al.: Nonrigid registration using
  free-form deformations: application to breast {MR} images. IEEE transactions
  on medical imaging  18(8),  712--721 (1999)

\bibitem{shen2005consistent}
Shen, D., Sundar, H., Xue, Z., Fan, Y., Litt, H.: Consistent estimation of
  cardiac motions by {4D} image registration. In: MICCAI. pp. 902--910 (2005)

\bibitem{shi2012comprehensive}
Shi, W., Zhuang, X., Wang, H., et~al.: A comprehensive cardiac motion
  estimation framework using both untagged and {3-D} tagged {MR} images based
  on nonrigid registration. IEEE transactions on medical imaging  31(6),
  1263--1275 (2012)

\bibitem{simonovsky2016deep}
Simonovsky, M., Guti{\'e}rrez-Becker, B., Mateus, D., et~al.: A deep metric for
  multimodal registration. In: MICCAI. pp. 10--18 (2016)

\bibitem{tobon2013benchmarking}
Tobon-Gomez, C., De~Craene, M., Mcleod, K., et~al.: Benchmarking framework for
  myocardial tracking and deformation algorithms: An open access database.
  Medical image analysis  17(6),  632--648 (2013)

\bibitem{tsai2016video}
Tsai, Y.H., Yang, M.H., Black, M.J.: Video segmentation via object flow. In:
  CVPR. pp. 3899--3908 (2016)

\bibitem{uzunova2017training}
Uzunova, H., Wilms, M., Handels, H., et~al.: Training {CNN}s for image
  registration from few samples with model-based data augmentation. In: MICCAI
  (2017)

\end{thebibliography}

\end{document}